# Suicide Phenotyping from Clinical Notes in Safety-Net Psychiatric Hospital Using Multi-Label Classification with Pre-Trained Language Models

Zehan Li, M.S[1], Yan Hu, M.S[1], Scott Lane, Ph.D[2], Salih Selek, M.D[2], Lokesh Shahani, M.D[2], Rodrigo Machado-Vieira, M.D, Ph.D[2], Jair Soares, M.D[2], Hua Xu, Ph. D[3], Hongfang Liu Ph. D[1,2], Ming Huang Ph. D[1]

[1]MacWilliam School of Biomedical Informatics, The University of Texas Health Science Center at Houston, TX, USA; [2]Department of Psychiatry & Behavioral Sciences, McGovern Medical School, The University of Texas Health Science at Houston, Houston, TX, USA; [3]Section of Biomedical Informatics and Data Science, School of Medicine, Yale University, New Haven, CT, USA

**Corresponding author:**
Ming Huang, PhD
Postal address: 7000 Fannin Street #Suite 600, Houston, TX 77030
E-mail: ming.huang@uth.tmc.edu
Telephone: 713-500-3900

**Abstract**

*Accurate identification and categorization of suicidal events can yield better suicide precautions, reducing operational burden, and improving care quality in high-acuity psychiatric settings. Pre-trained language models offer promise for identifying suicidality from unstructured clinical narratives. We evaluated the performance of four BERT-based models using two fine-tuning strategies (multiple single-label and single multi-label) for detecting coexisting suicidal events from 500 annotated psychiatric evaluation notes. The notes were labeled for suicidal ideation (SI), suicide attempts (SA), exposure to suicide (ES), and non-suicidal self-injury (NSSI). RoBERTa outperformed other models using multiple single-label classification strategy (acc=0.86, F1=0.78). MentalBERT (acc=0.83, F1=0.74) also exceeded BioClinicalBERT (acc=0.82, F1=0.72) which outperformed BERT (acc=0.80, F1=0.70). RoBERTa fine-tuned with single multi-label classification further improved the model performance (acc=0.88, F1=0.81). The findings highlight that the model optimization, pretraining with domain-relevant data, and the single multi-label classification strategy enhance the model performance of suicide phenotyping.*

**Introduction**

Suicide remains a leading cause of death and disability, posing a serious public health and clinical challenge worldwide. Each year, more than 720,000 people die by suicide globally, with over half a million lives lost in the U.S. from 2011 to 2022, culminating in a record 49,369 deaths in 2022[1,2]. The economic burden of suicide and non-suicidal self-injury (NSSI) in the U.S. is immense, with an estimated annual cost of $510 billion (by 2020) due to medical expenses, lost productivity, and reduced quality of life[3]. Beyond the financial toll, suicide deeply impacts families and communities, with exposure to suicide increasing the risk of further suicides within social networks[4,5]. For every completed suicide, an estimated 11 emergency department visits for self-injury, 52 attempts and 336 instances of serious suicidal thoughts occur, underscoring the continuous nature of suicidality and, at the same time, the significant opportunities for prevention[6].

Suicide is a complex phenomenon that cannot be adequately studied by examining a single category to reflect real-world conditions as it cannot factor in the fluid nature of suicidality. It often involves the coexistence of multiple stages, including suicidal ideation (SI), suicide attempts (SA), exposure to suicide (ES), and non-suicidal self-injury (NSSI). Individuals with psychiatric disorders face a particularly high risk, with studies showing that mental illness and substance abuse contribute to approximately 90% of suicide cases[7,8]. Many patients admitted to psychiatric services often present with SI in conjunction with a past or recent SA or NSSI[9]. According to the interpersonal theory of suicide, suicidality exists on a continuum from SI to completed suicide, and this continuum affects risk levels and treatment approaches[10]. Accurately detecting and categorizing these coexisting suicidal events and related factors is essential for implementing targeted interventions, allocating resources effectively, and improving clinical outcomes in a high acuity psychiatric care setting. Traditional methods relying on surveys often face limitations, as psychiatric patients may underreport or inaccurately recall suicidal thoughts and behaviors. Structured surveys are prone to recall bias and inconsistencies in participants' reports, creating bottlenecks at scale in both clinical and research settings[11,12]. The advent of electronic health records (EHRs) has opened new possibilities for studying documented suicidal behaviors, but inconsistencies in coding and under-documentation remain major obstacles, especially in low-resource institutions, such as safety-net hospitals[13]. One study shows that only 3% of patients with an indication of SI or SA in the notes field had corresponding diagnostic codes in EHRs, limiting the accuracy of predictive modeling methods and observational studies[14]. Suicidal events and related factors are more reliably captured in clinical narratives, such as initial psychiatric evaluation (IPE) notes, which provide a detailed account of patients' mental health status, life stressors, and behaviors, offering a more nuanced view of suicidality[15].

Given this rich source of unstructured data, leveraging Natural Language Processing (NLP) to extract instances of suicidality from psychiatric evaluation notes presents a promising opportunity to fill the reporting gap left by structured billing data[16]. Extensive work has been devoted to automatic detection and classification of suicidality using NLP in recent years[17-19]. As the field of artificial intelligence shifts towards deep learning (DL) approaches, more studies are employing DL models for suicide classification as a more advanced strategy. One study developed a text classification convolutional neural network (CNN) to detect suicide ideation from clinical notes. The CNN outperformed all other machine learning (ML) models tested in the study with the best performing F1 score of 0.82, achieving similar performance to predictive models in previous studies[17]. Another study also developed a CNN model that predicted suicide attempts from clinical notes better than ML models of F1 score of 0.92[19]. More recently, pretrained language models presented new opportunities to leverage the attention mechanism of the full text to improve NLP methods in predicting suicidal tendencies. One study developed a sentence level binary classification task based on RoBERTa-based models and achieved F1 score of 0.83 for either SI or SA and 0.78 in differentiating SI from SA (Bhanu Pratap Singh Rawat 2022), showcasing a reliable approach for fine-tuning pretrained language models with an additional annotated psychiatric corpus[18].

A common limitation of these studies, however, is their focus on binary classification tasks, where models either detect the presence of suicidality or differentiate between specific types (e.g., SI vs. SA)[18]. As demonstrated in one systematic review on the use of ML techniques for suicidal behavior prediction, all 35 included studies were implemented as binary classification tasks[20].

To the best of our knowledge, no previous studies have explored multi-label classification models for detecting multiple coexisting categories of suicidality within clinical notes. This research addresses the critical need for a system capable of automatically extracting large volumes of clinical notes in EHRs for various unstructured suicidal events and related factors. In this study, we developed and evaluated five multi-label classifiers based on four BERT-based language models, exploring the advantages of a single multi-label classification strategy over the traditional multiple single-label or binary classification approach. By comparing the classification performance of BERT-based language models (generic, domain-adapted, and disease-specific transformers) and evaluating two finetuning strategies (multiple single-label classification vs single multi-label classification), this study aims to advance the phenotyping performance of suicidality in psychiatric patients using pre-trained language models.

**Methods**

In the following sections, we will outline the steps taken for clinical dataset collection and analyis, including label characteristics and distribution. We will describe the selection process for the four pre-trained language models and two classification strategies adopted. Additionally, we will detail the finetuning design and conclude with the evaluation of model performance at both model and single label levels. (Figure 1).

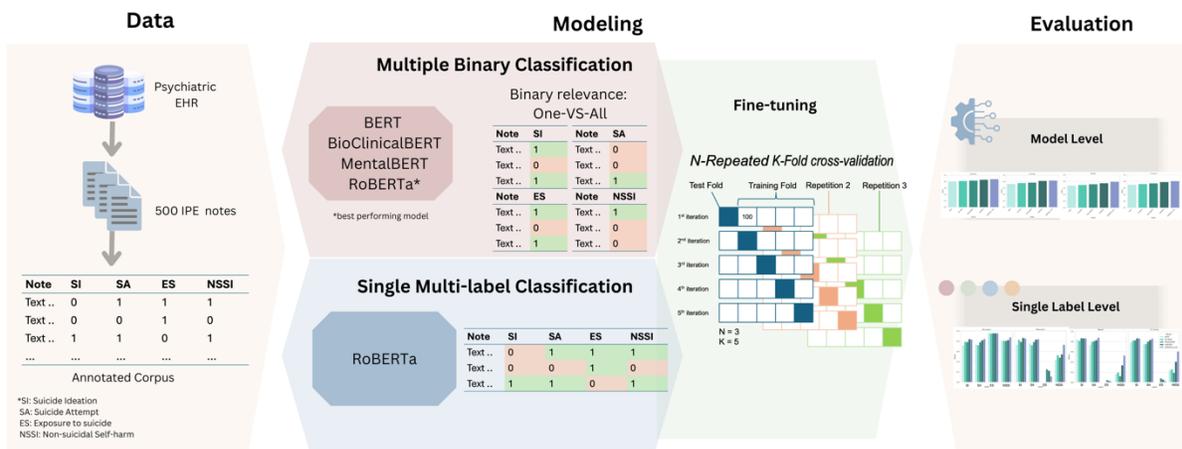

**Figure 1.** Graphical abstract for suicide phenotyping with pre-trained language models

**Clinical dataset**

The experiments and findings of this study are based on 500 IPE notes collected from the EHR system at Harris County Psychiatric Center (HCPC), a safety-net psychiatric hospital affiliated with the University of Texas Health Science Center at Houston (UTHealth). This study was approved by the Institutional Review Board (IRB# HSC-SBMI-17-0354) at UTHealth. The collected IPE notes span from 2001 to 2021 and represent a diverse range of patients receiving inpatient psychiatric care. Our research team, including a panel of psychiatrists, developed the annotation guideline and annotated the IPE notes . The notes were labeled into four suicide-related categories: Suicidal Ideation (SI), Suicide Attempt (SA), Exposure to Suicide (ES), and Non-Suicidal Self-Injury (NSSI). SI includes mentions of thoughts or plans to kill oneself or suicide, while SA refers to actual attempts to end one's life. ES captures experiences of individuals exposed to the suicide of others, and NSSI includes deliberate self-harm without the intention to cause death.

The distribution of each label is shown in Figure 2. The 500 notes were assigned with a total of 675 labels, specifically 294 SI, 265 SA, 22 ES, and 94 NSSI. Of the 500 notes, 103 are free from any suicide mentions. A total of 172 notes (34.4%) contained only one label, with the majority being SI (N=96) or SA (N=62), while smaller counts included NSSI (N=11) and ES (N=3). Conversely, 225 notes (45%) contained more than one label, with the most common combination being SI and SA (N=178, 35.6%). Furthermore, a subset of 45 notes (9%) contained three labels, while 4 notes (0.8%) contained all four labels (SI, SA, ES, and NSSI). This distribution reflects the high acuity and complexity of suicidal behaviors in the inpatient psychiatric population at HCPC, highlighting the coexisting nature of suicidal events and related factors.

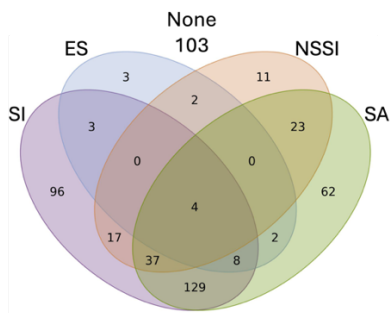

**Figure 2**. Label distribution of co-existing suicidal events and related factors. SI: Suicide Ideation; SA: Suicide Attempt; ES: Exposure to Suicide; NSSI: Non-Suicidal Self-Injury

**Pre-trained language models**

Pre-trained language models, based on the transformer architecture, are trained on large text corpora using self-supervised learning techniques, where models learn to predict missing or masked tokens within sentences. This pretraining enables the models to capture rich linguistic representations, including syntax and semantics. These models can be further fine-tuned on specific downstream tasks, making them highly effective for a wide range of NLP applications. In this study, We employed four widely used pre-trained language models based on BERT (Bidirectional Encoder Representations from Transformers) to build text classifiers to identify suicidal events and factors in IPE notes and evaluate their performance[21]. These BERT models are categorized into three types: generic models, domain-adapted models, and disease-specific models.

*Generic models*

The generic models used in this study include BERT ("bert-base-uncased") and RoBERTa ("roberta-base")[22]. BERT is a pre-trained language model developed by Google that utilizes a transformer architecture with a self-attention mechanism, allowing it to learn word and phrase relationships from large datasets. It has gained widespread popularity for its strong performance across NLP tasks. RoBERTa, based on the same architecture as BERT, is developed with more robust training methodologies. Both models were trained using book corpus and English Wikipedia, which are not tied to any specific domain knowledge. Compared with BERT, RoBERTa leverages a larger and more diverse dataset, extends the training duration, and utilizes dynamic token masking, and omits the Next Sentence Prediction task included in BERT.

*Domain-adapted model*

We utilized BioClinicalBERT ("emilyalsentzer/BioClinicalBERT"), a domain-adapted model specifically designed for NLP tasks in the biomedical and clinical domain[23]. BioClinicalBERT builds on the foundation of BioBERT, an extension of the original BERT model that was further pre-trained on biomedical literature to enhance performance on medical and biological texts. BioClinicalBERT goes beyond this by incorporating training data from MIMIC-III, a publicly available database containing EHRs from Intensive Care Unit (ICU) patients. The MIMIC-III dataset includes a wide range of clinical documents, such as discharge summaries, nursing notes, and physician reports, allowing BioClinicalBERT to capture the unique language, terminology, and structure prevalent in clinical documentation. We employed BioClinicalBERT in this study in order to examine its BioClinicalBERT's capability for clinical NLP tasks that involve identifying meaningful insights from unstructured clinical data.

*Disease-specific model*

We also employed MentalBERT, a disease-specific language model tailored for mental healthcare applications[24]. MentalBERT was developed to address the gap in domain-specific models for mental health by pretraining on mental health-related text, primarily collected from a social forum - Reddit. The pretraining corpus includes discussions from subreddits focused on mental health issues such as depression, anxiety, and suicidal ideation, including communities like "r/depression," "r/SuicideWatch," and "r/mentalhealth." The use of MentalBERT in this study aims to evaluate its performance to identify suicide-related behaviors documented in clinical notes, compared with traditional language models without exposure to the specialized vocabulary and mental health context.

**Multi-label classification**

Multi-label classification is a classification paradigm in which each instance can be associated with multiple labels or categories simultaneously, rather than being restricted to a single label. This approach enables the prediction of multiple labels for each instance, better capturing the complexity of real-world scenarios where instances frequently belong to more than one category. Multi-label classification is particularly suited for tasks where objects or data points exhibit multiple attributes or characteristics that cannot be encapsulated by a single label. Multi-label classification techniques can be broadly categorized into two major approaches -- problem transformation and algorithm adaptation. Problem transformation methods transform a multi-label classification problem into multiple single-label problems, enabling the use of traditional single-label classifiers to address the task. In contrast, algorithm adaptation methods extend existing learning algorithms to directly handle multi-label data, allowing for the simultaneous prediction of multiple labels. In this study, we explored one popular problem transformation method, Binary Relevance Approach, as well as an algorithm adaptation method, the multi-label neural network.

We utilized a One-vs-All Binary relevance method to convert a multi-label dataset into multiple single-label binary datasets. Specifically, this method transforms the dataset with k labels into k single-label datasets, with a binary classifier developed for each label. In our case, this method transforms our dataset with four labels into four separate single-label datasets as shown in Table 1. Four binary classifiers were then developed to identify each of four suicide-related events and factors (SI, SA, ES, or NSSI) by using pretrained language models.

**Table 1.** Sample datasets with annotation for each label for binary classification

| Note   | SI | Note   | SA | Note   | ES | Note   | NSSI |
|--------|----|--------|----|--------|----|--------|------|
| Text.. | 1  | Text.. | 0  | Text.. | 1  | Text.. | 1    |
| Text.. | 0  | Text.. | 0  | Text.. | 0  | Text.. | 0    |
| Text.. | 1  | Text.. | 1  | Text.. | 1  | Text.. | 0    |

We employed the multi-label neural network approach to implement multi-label classification for detecting co-existing suicide events and factors in the notes with pre-trained language models. In the classification layer, we applied a separate sigmoid activation function, as opposed to a SoftMax function used in multi-class classification, for the prediction for each label. The sigmoid function produces a probability for each label, reflecting the likelihood that the label is relevant to the input. By assigning a distinct sigmoid function to each label, we convert the logits into probabilities, enabling the model to predict multiple labels concurrently while accounting for inter-label relationships.

**Model training and Evaluation**

In this study, we implemented multi-label classification to identify coexisting suicidal events and related factors at the document level through multiple binary classification models and single multi-label classification model. The multiple binary classification strategy was applied to the four pre-trained language models: BERT, RoBERTa, BioClinicalBERT, and MentalBERT. The best-performing model from the multiple binary classification was selected to implement the single multi-label classification method. We fine-tuned each classification model, either binary classification model or multi-label classification model, on a training set of 400 IPE notes and evaluated their performance on a test set of 100 notes. To enhance the reliability of our results and reduce sampling bias, we used Repeated Stratified K-Fold cross-validation. The training and evaluation were performed using 5-fold cross-validation with 3 repetitions.

The maximum input length was set to 512 tokens, with longer texts truncated as needed. Notably, 431 (86.2%) of the notes had less than 512 tokens. For binary classification task, the training process used the following hyperparameters: a learning rate of 1e5, a batch size of 4, and 5 training epochs, with a weight decay of 0.01 using the AdamW optimizer. The multi-label classification model was trained with a learning rate of 2e-5, a batch size of 8, 20 training epochs, and a weight decay of 0.01. Different training epochs and learning rates were employed for the binary and multi-label classification tasks to ensure proper convergence of pre-trained language models.

To evaluate model performance, we calculated accuracy, precision, recall, and F1 scores at both label and model levels. The F1 score, being the harmonic mean of precision and recall, is especially valuable in scenarios with imbalanced labels, as it provides a balanced measure of model performance. We computed accuracy, precision, recall, and F1 scores for each single label to measure the performance at the label level. Furthermore, we reported overall accuracy and micro-average precision, recall, and F1 scores to provide an overview of model performance. All metrics were averaged over five cross-validation folds and three repetitions to ensure robust evaluation. We calculated standard deviations to provide insights into the variability of the model's performance across different experiments.

**Result**

We assessed classification performance of four pre-trained language models (BERT, RoBERTa, BioClinicalBERT, and MentalBERT) and two multi-label classification strategies (RoBERTa vs RoBERTa_multi) at two levels: the overall model level and the individual label level. (Figure 3). RoBERTa_multi denotes RoBERTa used in a single multi-label classification.

*Model level*

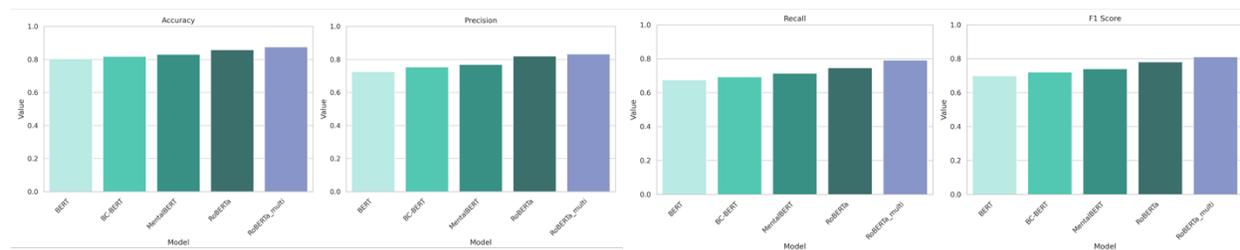

**Figure 3.** Model level performance. BERT, BC_BERT(BioClinicalBERT), MentalBERT and RoBERTa are pre-trained language models used for multiple binary classification. RoBERTa_multi denotes RoBERTa used in a single multi-label classification.

In the multiple single-label classification, RoBERTa achieved the best performance with a micro-average accuracy of 0.86 ± 0.01 and F1 score of 0.78 ± 0.01. Compared to the BERT model, which achieved an accuracy of 0.80 ± 0.01 and an F1 score of 0.70 ± 0.02, RoBERTa demonstrated superior performance. MentalBERT, a disease-specific model pre-trained on mental health-related text from social media platforms, achieved an accuracy of 0.83 ± 0.01 and F1 score of 0.74 ± 0.01, outperforming BioClinicalBERT, a health domain-adapted model, which obtained an accuracy of 0.82 ± 0.01 and F1 score of 0.72 ± 0.01.

Given the superior performance of RoBERTa in binary classification for individual labels, we selected RoBERTa to train a single multi-label classification model using the multi-label neural network approach. This resulted in a 4% improvement in performance compared to RoBERTa trained with binary relevance method, achieving an accuracy of 0.88 ± 0.01 and an F1 score of 0.81 ± 0.01.

*Single label level*

The evaluation of performance for each of the four suicidal events and related factors is detailed in Figure 4. Each label was assessed independently to understand how specific label characteristics influenced model performance. Across all five models tested, the best performances were consistently observed for the SI and SA labels, with accuracy ranging from 0.73 to 0.85 and F1 scores between 0.73 and 0.86.

While BERT-based models demonstrated strong accuracy, they exhibited lower F1 scores for ES and NSSI, largely due to the smaller number of positive cases for these labels. This label imbalance significantly impacted the models' performance. Among the models, RoBERTa consistently outperformed other BERT variants, achieving F1 scores of 0.86 for SI, 0.83 for SA, 0.05 for ES, and 0.40 for NSSI. Except ES label, this result highlights RoBERTa's robustness across all categories, particularly its ability to handle the more imbalanced labels like NSSI.

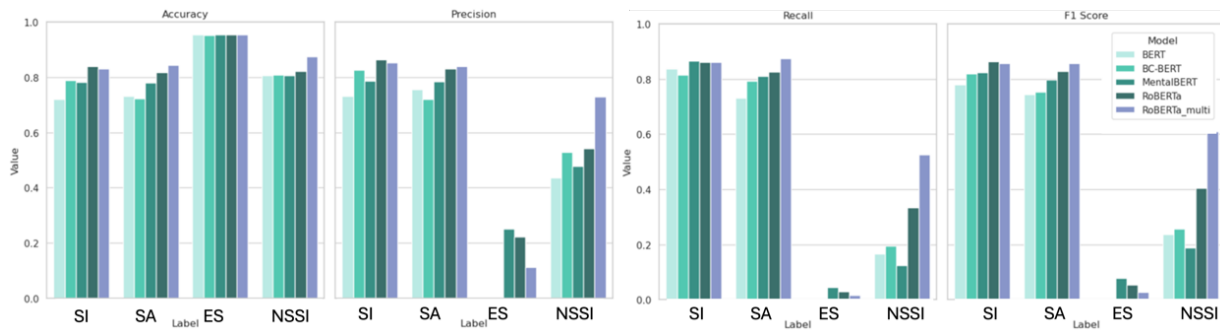

**Figure 4.** Label Level performance. BERT, BC_BERT(BioClinicalBERT), MentalBERT and RoBERTa are pre-trained language models used for multiple binary classification. RoBERTa_multi denotes RoBERTa used in a single multi-label classification. SI: Suicide Ideation; SA: Suicide Attempt; ES: Exposure to Suicide; NSSI: Non-Suicidal Self-Injury.

BERT and BioClinicalBERT failed to learn from the data for the ES label, resulting in precision and recall scores of zero. In contrast, the RoBERTa multiple single-label classifier learned from all labels successfully, achieving the highest overall F1 score of 0.86 ± 0.01. The RoBERTa single multi-label classifier also improved performance on SA with an F1 score of 0.86 ± 0.01 and showed notable improvement on NSSI with a F1 of 0.61 ± 0.03. This suggests that the single multi-label classifier is a more effective approach for clinical classification tasks, although challenges persist with less frequent labels such as ES.

**Discussion**

In this study, we developed and evaluated five multi-label classifiers based on four BERT-based language models, exploring the advantages of a single multi-label classification strategy over the traditional multiple single-label classification approach. Our findings revealed several key insights, particularly in terms of prediction difficulty across labels, model performance, and the benefits of multi-label classification.

The corpus used to pre-train the generic BERT model did not include domain-specific knowledge, which contributed to their lower performance compared to domain-specific models. BioClinicalBERT, pre-trained on medical records, and MentalBERT, pre-trained on mental health-related text from social media platforms, were able to leverage domain-specific insights to achieve better results. Consistent with prior study, our finding indicated that disease-specific and domain-adapted models such as MentalBERT and BioClinicalBERT showed improvements over generic BERT model[25]. MentalBERT performed better than BioClinicalBERT due to its more relevant pretrained corpus with a special focus on mental health, including posts about suicide ideation and conditions (e.g. depression and anxiety). These models were more adept at understanding the specialized language and context surrounding mental health and suicidality. Among all the tested models, RoBERTa consistently delivered the best results overall. RoBERTa's superior performance can be attributed to its dynamic batch training and extensive exposure to a broader training corpus, which allowed it to better capture the linguistic patterns relevant to mentions of suicidal events and related factors.

With the highly imbalanced dataset, SI and SA are good due to the larger number of positive cases, both over 200. All models had lower performance on ES (21 cases) and NSSI (94 cases). After straining split, each fold contains less than 16 positive cases of ES. Error analysis also highlighted the challenges in predicting certain labels, particularly ES and NSSI. These labels, characterized by fewer positive examples in the dataset, proved difficult for BERT-based models to accurately detect. For instance, subtle mentions of ES, such as "friend attempted suicide" or "witnessed family suicide," were often missed due to the scarcity of training data after stratification. Similarly, indirect references to NSSI (e.g., "self-harm without intent") were harder for the models to capture with high precision. While BERT, BioClinicalBERT, and MentalBERT struggled to accurately classify NSSI, RoBERTa demonstrated significant improvement in performance. Especially, RoBERTa single multi-label classification classifier improved additional 20% in NSSI label compared to its counterpart trained with multiple single-label classification approach.

Furthermore, our study demonstrated that single multi-label classification strategy was superior to multiple single-label strategies. RoBERTa's multi-label classifier provided an efficient and effective way to predict multiple labels simultaneously, reducing the number of models needed for training while capturing the interrelationships between labels. This approach is particularly valuable in clinical settings where co-occurring suicidal events and related factors, such as SI and SA, are common. The ability of a single multi-label classifier to learn the relationships between labels not only improves performance but also makes it a more cost-effective solution for clinical applications. The co-existence of multiple suicide-related factors identified with a single model could lead to more accurate and comprehensive risk assessments, ultimately improving decision-making processes in healthcare environments. By understanding how these factors overlap, clinicians can develop more targeted interventions for individuals at risk, thereby improving patient outcomes while reducing healthcare resource use.

*Limitations and Future Steps*

The clinical notes were collected from HCPC, an inpatient psychiatric hospital at UTHealth, and may not be representative of patient populations or clinical settings in other regions of the country. The relatively small sample size and limited label diversity likely affected the models' ability to generalize, particularly for underrepresented labels like ES and NSSI. Future research could benefit from using larger datasets with more diverse labels. Weak labeling techniques could be employed to generate "silver-standard" datasets for training, which would improve model robustness[26]. We could also further explore data augmentation methods through generative AI models to create synthetic data for model training. We did not implement generative AI models, such as ChatGPT, in the current study for either data augmentation or multi-label classification due to HIPAA and privacy concerns surrounding the use of clinical data on third-party servers. Future work could explore advanced open-sourced LLMs, such as Llama3.1, with a focus on overcoming these privacy challenges.

**Conclusion**

To further our understanding of the suicide continuum and to generate real-world evidence using unstructured EHR data from a safety-net psychiatric hospital, we explored multi-label classification with pre-trained language models for detecting multiple coexisting categories of suicidality within psychiatric evaluation notes. Our finding is consistent with prior study that disease-specific and domain-adapted models such as MentalBERT and BioClinicalBERT showed improvements over generic BERT model[25]. The main contribution of this study is to demonstrate that single multi-label classification offers significant advantages over traditional multiple single-label or binary classification methods, particularly in capturing the co-occurrence of suicidal ideation, attempts, and non-suicidal self-injury. This finding shed lights for developing targeted interventions, improving resource allocation, and enhancing patient care in high-acuity psychiatric settings. Moving forward, advanced NLP techniques with generative AI models and larger datasets should be adopted to further refine suicide phenotyping models, ultimately contributing to improved clinical decision-making and patient outcomes.

**Acknowledgement**

This research work was supported by the National Library of Medicine under award number R01-LM011934 at National Institutes of Health. The content is solely the responsibility of the authors and does not necessarily represent the official views of the National Institutes of Health.